%% file: main.tex
\documentclass[conference]{IEEEtran}
\IEEEoverridecommandlockouts
\usepackage{cite}
\usepackage{amsmath,amssymb,amsfonts}
\usepackage{algorithmic}
\usepackage{graphicx}
\usepackage{textcomp}
\usepackage{xcolor}
\usepackage{algorithm}
\usepackage{algorithmic}
\usepackage{multirow}
\usepackage{subfigure}
\usepackage{setspace}
\usepackage{amsthm}
\usepackage{url}
\def\BibTeX{{\rm B\kern-.05em{\sc i\kern-.025em b}\kern-.08em
    T\kern-.1667em\lower.7ex\hbox{E}\kern-.125emX}}
\begin{document}

\title{\textsc{HiGitClass}: Keyword-Driven Hierarchical Classification of GitHub Repositories
}

\author{
\IEEEauthorblockN{Yu Zhang$^1$, Frank F. Xu$^2$, Sha Li$^1$, Yu Meng$^1$, Xuan Wang$^1$, Qi Li$^3$, Jiawei Han$^1$}
\IEEEauthorblockA{$^1$Department of Computer Science, University of Illinois at Urbana-Champaign, Urbana, IL, USA \\
$^2$Language Technologies Institute, Carnegie Mellon University, Pittsburgh, PA, USA \\
$^3$Department of Computer Science, Iowa State University, Ames, IA, USA \\
\{yuz9, shal2, yumeng5, xwang174, hanj\}@illinois.edu, \ \ frankxu@cmu.edu, \ \ qli@iastate.edu}
}

\maketitle

\begin{spacing}{0.99}
\begin{abstract}
    \input{0-abstract.tex}
\end{abstract}

\begin{IEEEkeywords}
hierarchical classification, GitHub, weakly-supervised learning
\end{IEEEkeywords}

\input{1-intro.tex}

\input{2-prelim.tex}
\input{3-method.tex}
\input{4-exp.tex}
\input{5-related.tex}
\input{6-conclusion.tex}

\section*{Acknowledgment}
We thank Xiaotao Gu for useful discussions. The research was sponsored in part by U.S. Army Research Lab. under Cooperative Agreement No. W911NF-09-2-0053 (NSCTA), DARPA under Agreement No. W911NF-17-C0099, National Science Foundation IIS 16-18481, IIS 17-04532, and IIS-17-41317, DTRA HDTRA11810026, and grant 1U54GM114838 awarded by NIGMS through funds provided by the trans-NIH Big Data to Knowledge (BD2K) initiative (www.bd2k.nih.gov).  
The views and conclusions contained in this paper are those of the authors and should not be interpreted as representing any funding agencies.
We thank anonymous reviewers for valuable and insightful feedback.

\bibliographystyle{abbrv}
\bibliography{icdm19}

\end{spacing}

\end{document}

%% file: 0-abstract.tex
GitHub has become an important platform for code sharing and scientific exchange. With the massive number of repositories available, there is a pressing need for topic-based search. Even though the topic label functionality has been introduced, the majority of GitHub repositories do not have any labels, impeding the utility of search and topic-based analysis. 
This work targets the automatic repository classification problem as \textit{keyword-driven hierarchical classification}. Specifically, users only need to provide a label hierarchy with keywords to supply as supervision.
This setting is flexible, adaptive to the users' needs, accounts for the different granularity of topic labels and requires minimal human effort.
We identify three key challenges of this problem, namely (1) the presence of multi-modal signals; (2) supervision scarcity and bias; (3) supervision format mismatch.
In recognition of these challenges, we propose the \textsc{HiGitClass} framework, comprising of three modules: heterogeneous information network embedding;  keyword enrichment; topic modeling and pseudo document generation. 
Experimental results on two GitHub repository collections confirm that \textsc{HiGitClass} is superior to existing weakly-supervised and dataless hierarchical classification methods, especially in its ability to integrate both structured and unstructured data for repository classification.

%% file: 1-intro.tex
\section{Introduction}
For the computer science field, code repositories are an indispensable part of the knowledge dissemination process, containing valuable details for reproduction. For software engineers, sharing code also promotes the adoption of best practices and accelerates code development. 
The needs of the scientific community and that of software developers have facilitated the growth of online code collaboration platforms, the most popular of which is GitHub, with over 96 million repositories and 31 million users as of 2018. 
With the overwhelming number of repositories hosted on GitHub, there is a natural need to enable search functionality so that users can quickly target repositories of interest. 
To accommodate this need, GitHub introduced topic labels\footnote{\url{https://help.github.com/en/articles/about-topics}} which allowed users to declare topics for their own repositories. However, topic-based search on GitHub is still far from ideal. For example, when searching for repositories related to ``phylogenetics'', 
a highly relevant repository \texttt{opentree}\footnote{\url{https://github.com/OpenTreeOfLife/opentree}} with many stars and forks
does not even show up in the first 10 pages of search results as it does not contain the ``phylogenetics'' tag. Hence, to improve the search and analysis of GitHub repositories, a critical first step is \textit{automatic repository classification}. 

In the process of examining the automatic repository classification task, we identify three different cases of missing labels: (1) \textit{Missing annotation}: the majority of repositories (73\% in our \textsc{Machine-Learning} dataset and 78\% in our \textsc{Bioinformatics} dataset) have no topic labels at all; (2) \textit{Incomplete annotation}: since topic labels can be arbitrarily general or specific, some repositories may miss coarse-grained labels while others miss fine-grained ones; (3) \textit{Evolving label space}: related GitHub topics tags may not have existed at the time of creation, so the label is naturally missing. Missing annotation is the major drive behind automatic classification, but this also implies that labeled data is scarce and expensive to obtain. 
Incomplete annotation reflects the hierarchical relationship between labels: repositories should not only be assigned to labels of one level of granularity, but correspond to a path in the class hierarchy.
Finally, the evolving label space requires the classification algorithm to quickly adapt to a new label space, or take the label space as part of the input. 
Combining these observations, we define our task as \textit{keyword-driven hierarchical classification} for GitHub repositories. By keyword-driven, we imply that we are performing classification using only a few keywords as supervision. 

Compared to the common setting of fully-supervised classification of text documents, \textit{keyword-driven hierarchical classification} of GitHub repositories poses unique challenges. 
First of all, GitHub repositories are complex objects with metadata, user interaction and textual description.  
As a result, multi-modal signals can be utilized for topic classification, including user ownership information, existing tags and README text. To jointly model structured and unstructured data, we propose to construct a \textit{heterogeneous information network} (HIN) centered upon words. By learning node embeddings in this HIN, we obtain word representations that reflect the co-occurrence of multi-modal signals that are unique to the GitHub repository dataset. 
We also face the supervision scarcity and bias problem as users only provide one keyword for each class as guidance. This single keyword may reflect user's partial knowledge of the class and may not achieve good coverage of the class distribution. 
In face of this challenge, we introduce a \textit{keyword enrichment} module that 
expands the single keyword to a keyword \textit{set} for each category.
The newly selected keywords are required to be 
close to the target class in the embedding space. Meanwhile, we keep mutual exclusivity among keyword sets so as to create a clear separation boundary. 
Finally, while users provide a label hierarchy, the classification algorithm ultimately operates on repositories, so there is a mismatch in the form of supervision. Since we already encode the structured information through the HIN embeddings, in our final classification stage we represent each repository as a document. 
To transform keywords into documents, we first model each class as a topic distribution over words and estimate the distribution parameters. Then based on the topic distributions, we follow a two-step procedure to generate pseudo documents for training. 
This also allows us to employ powerful classifiers such as CNNs for classification, which would not be possible with the scarce labels.

To summarize, we have the following contributions:
\begin{itemize}
    \item We present the task of keyword-driven hierarchical classification of GitHub repositories. While GitHub has been of widespread interest to the research community, no previous efforts have been devoted to the task of automatically assigning topic labels to repositories, which can greatly facilitate repository search and analysis. To deal with the evolving hierarchical label space and circumvent expensive annotation efforts, we only rely on the user-provided label hierarchy and keywords to train the classifier.
    \item We design the \textsc{HiGitClass} framework, which consists of three modules: HIN construction and embedding; keyword enrichment; topic modeling and pseudo document generation. The three modules are carefully devised to overcome three identified challenges of our problem: the presence of multi-modal signals; supervision scarcity and bias; supervision format mismatch. 
    \item We collect two datasets of GitHub repositories from the machine learning and bioinformatics research community. On both datasets we show that our proposed framework \textsc{HiGitClass} outperforms existing supervised and semi-supervised models for hierarchical classification.
\end{itemize}
The remainder of this paper is organized as follows.
In Section II, we formally introduce our problem definition, multi-modal signals in a GitHub repository and heterogeneous information networks. 
In Section III, we elaborate our framework \textsc{HiGitClass} with its three components. 
Then in Section IV, we present experimental results and discuss our findings. Section V covers related literature and we conclude in Section VI.

%% file: 2-prelim.tex
\section{Preliminaries}
\subsection{Problem Definition}
We study hierarchical classification of GitHub repositories where the categories form a tree structure. Traditional hierarchical classification approaches \cite{dumais2000hierarchical,liu2005support} rely on a large set of labeled training documents. In contrast, to tackle the evolving label space and alleviate annotation efforts, we formulate our task as \textit{keyword-driven} classification, where users just need to provide the label hierarchy and \textit{one} keyword for each \textit{leaf} category. This bears some similarities with the dataless classification proposed in \cite{song2014dataless} that utilizes an user-defined label hierarchy and class descriptions. There is no requirement of \textit{any} labeled repository. Formally, our task is defined as follows.

\vspace{1mm}

\noindent \textbf{Problem Definition.} (\textsc{Keyword-Driven Hierarchical Classification}.)
\textit{Given a collection of unlabeled GitHub repositories, a tree-structured label hierarchy $\mathcal{T}$ and one keyword $w_{i0}$ for each leaf class $C_i$ $(i=1,...,\mathcal{L})$, our task is to assign appropriate category labels to the repositories, where the labels can be either a leaf or an internal node in $\mathcal{T}$.}

\subsection{GitHub Repositories}
\begin{figure}[t]
\centering
\includegraphics[width=0.48\textwidth]{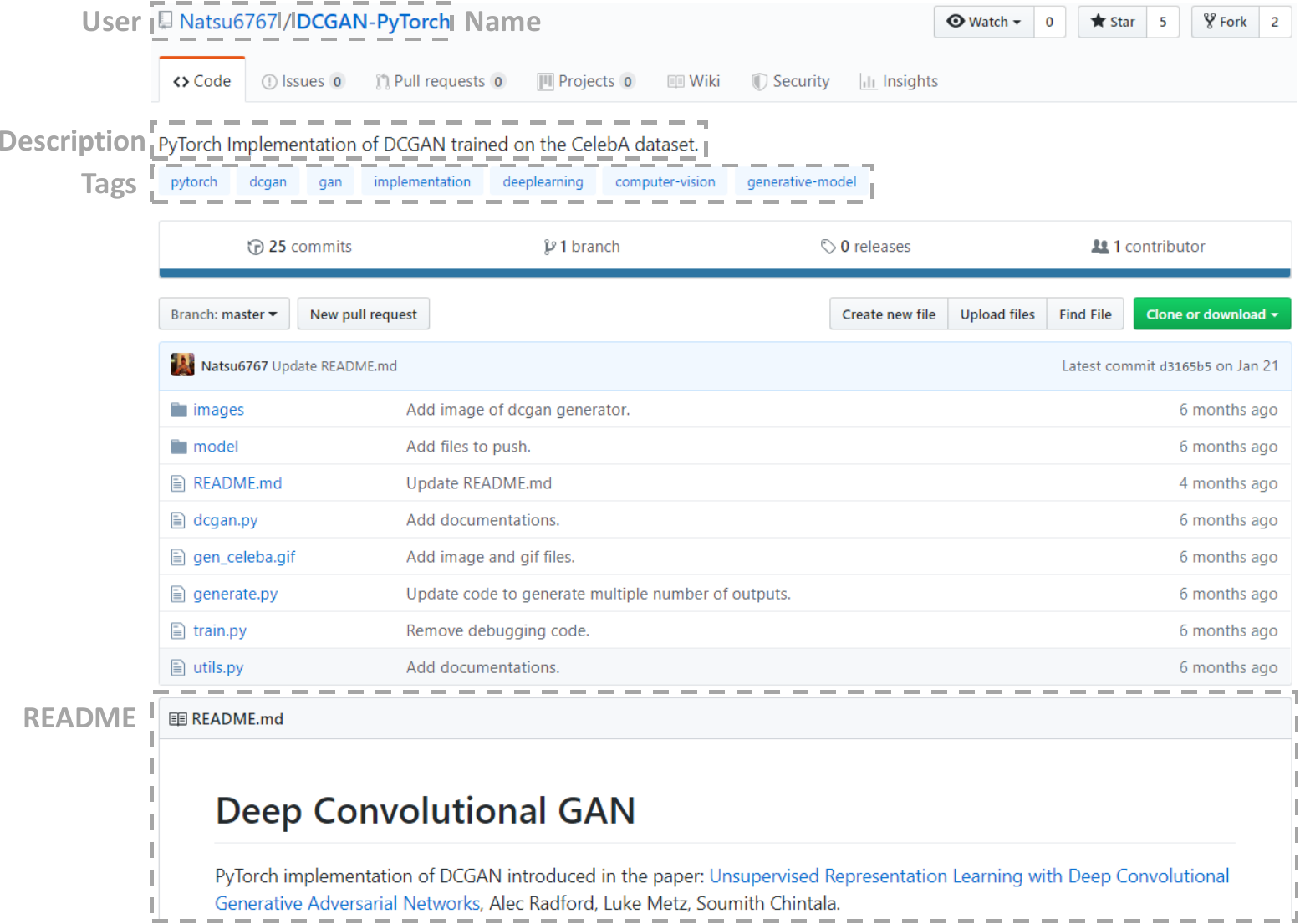}
\caption{A sample GitHub repository with the user's name, the repository name, description tags, and README (only the first paragraph is shown).} \label{fig:repo}
\vspace{-1em}
\end{figure}

\begin{figure*}[t]
\centering
\includegraphics[width=0.95\textwidth]{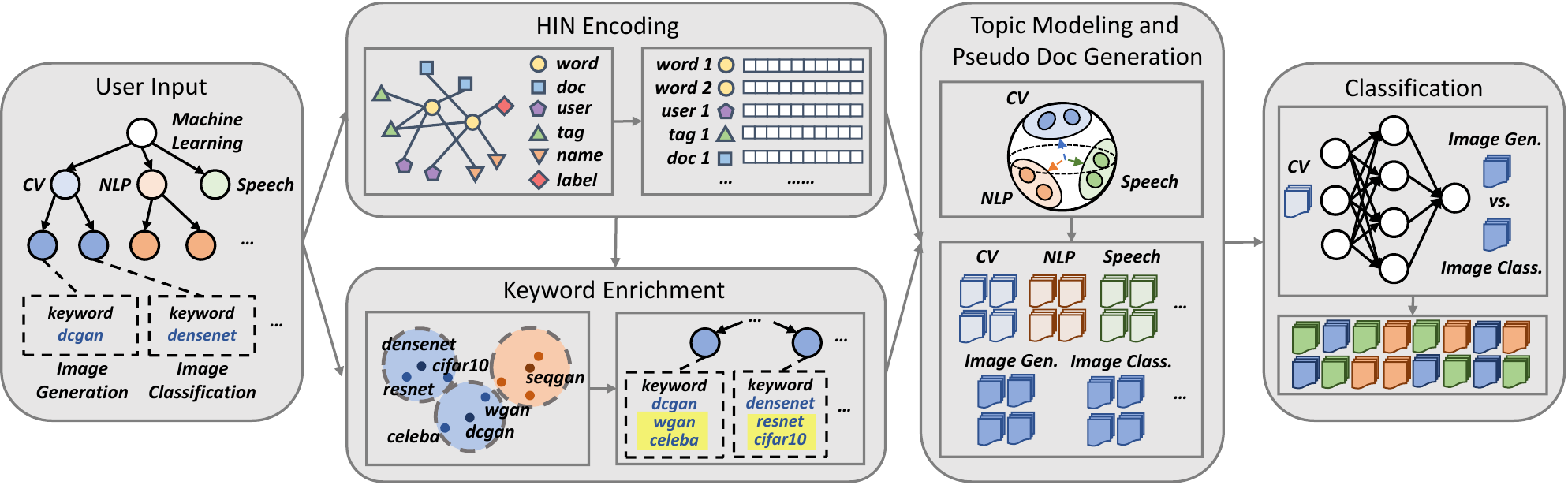}
\caption{The \textsc{HiGitClass} framework. Three key modules (i.e., HIN encoding, keyword enrichment, and pseudo document generation) are used to tackle the aforementioned three challenges, respectively.} \label{fig:framework}
\vspace{-0.5em}
\end{figure*}

Fig. \ref{fig:repo} shows a sample GitHub repository\footnote{https://github.com/Natsu6767/DCGAN-PyTorch}.
With the help of GitHub API\footnote{https://developer.github.com/v3/}, we are able to extract comprehensive information of a repository including metadata, source code and team dynamics. In \textsc{HiGitClass}, we utilize the following information:

\vspace{1mm}

\noindent \textbf{User.} Users usually have consistent interests and skills. If two repositories share the same user (``\textit{Natsu6767}'' in Fig. \ref{fig:repo}), they are more likely to have similar topics (e.g., deep learning or image generation). 

\vspace{1mm}

\noindent \textbf{Name.} If two repositories share the same name, it is likely that one is forked from the other and they should belong to the same topic category. Besides, indicative keywords can be obtained by segmenting the repository name properly (e.g., ``\textit{DCGAN}'' and ``\textit{PyTorch}'' in Fig. \ref{fig:repo}).

\vspace{1mm}

\noindent \textbf{Description.} The description is a concise summary of the repository. It usually contains topic-indicating words (e.g., ``\textit{DCGAN}'' and ``\textit{CelebA}'' in Fig. \ref{fig:repo}). 

\vspace{1mm}

\noindent \textbf{Tags.} Although a large proportion of GitHub repositories are not tagged, when available, tags are strong indicators of a repository's topic (e.g., ``\textit{dcgan}'' and ``\textit{generative-model}'' in Fig. \ref{fig:repo}). 

\vspace{1mm}

\noindent \textbf{README.} The README file is the main source of textual information  in a repository. In contrast to the description, it elaborates more on the topic but may also diverge to other issues (e.g., installation processes and code usages). The latter introduces noises to the task of topic inference. 

We concatenate the description and README fields into a single \textbf{Document} field, which serves as the textual feature of a repository.

\subsection{Heterogeneous Information Networks}
In our proposed framework \textsc{HiGitClass}, we model the multi-modal signals of GitHub repositories as a heterogeneous information network (HIN)~\cite{sun2011pathsim,sun2012mining}. HINs are an extension of homogeneous information networks to support multiple node types and edge types. We formally define a heterogeneous information network as below:


\vspace{1mm}

\noindent \textbf{Heterogeneous Information Network (HIN).} \textit{An HIN is defined as a graph $G = (\mathcal{V}, \mathcal{E})$ with a node type mapping $\phi: \mathcal{V}\rightarrow\mathcal{T}_\mathcal{V}$ and an edge type mapping $\psi: \mathcal{E}\rightarrow\mathcal{T}_\mathcal{E}$. Either the number of node types $|\mathcal{T}_\mathcal{V}|$ or the number of relation types $|\mathcal{T}_\mathcal{E}|$ is larger than 1.}

As we all know, one advantage of networks is the ability to go beyond direct links and model higher-order relationships which can be captured by paths between nodes. We introduce the notion of meta-paths, which account for different edge types in HINs. 

\vspace{1mm}

\noindent \textbf{Meta-Path.}  \textit{In an HIN, meta-paths \cite{sun2011pathsim} are an abstraction of paths proposed to describe multi-hop relationships.  For an HIN $G = (\mathcal{V}, \mathcal{E})$, a \textit{meta-path} is a sequence of edge types $\mathcal{M} = E_1$-$E_2$-...-$E_L$ ($ E_i \in \mathcal{T}_\mathcal{E}$). Any path that has the same types as the meta-path is an instance of the meta-path. When edge types are a function of the node types, we also represent a meta-path as 
$\mathcal{M} = V_1$-$V_2$-...-$V_L$ ($V_i \in \mathcal{T}_\mathcal{V}$ and $V_i$-$V_{i+1} \in \mathcal{T}_\mathcal{E}$ for any $i$).} 

%% file: 3-method.tex
\section{Method}

We lay out our \textsc{HiGitClass} framework in Fig. \ref{fig:framework}. \textsc{HiGitClass} consists of three key modules, which are proposed to solve the three challenges mentioned in Introduction, respectively. 

To deal with \textit{multi-modal signals}, we propose an \textit{HIN encoding} module (Section \ref{sec:encode}). Given the label hierarchy and keywords, we first construct an HIN to characterize different kinds of connections between words, documents, users, tags, repository names and labels. Then we adopt \textsc{ESim} \cite{shang2016meta}, a meta-path guided heterogeneous network embedding technique, to obtain good node representations. 

To tackle \textit{supervision scarcity and bias}, we introduce a \textit{keyword enrichment} module (Section \ref{sec:enrich}). This module 
expands the user-specified keyword to a semantically concentrated keyword set for each category. The enriched keywords are required to 
share high proximity with the user-given one from the view of embeddings. 
Meanwhile, we keep mutual exclusivity among keyword sets so as to create a clear separation boundary.

To overcome \textit{supervision format mismatch}, we present a \textit{pseudo-document generation} technique (Section \ref{sec:gen}). We first model each class as a topic distribution over words and estimate the distribution parameters. Then based on the topic distributions, we follow a two-step procedure to generate pseudo documents for training. This step allows us to employ powerful classifiers such as \textit{convolutional neural network} \cite{kim2014convolutional}. Intuitively, the neural classifier is fitting the learned word distributions instead of a small set of keywords, which can effectively prevent it from overfitting.

\subsection{HIN Construction and Embedding}
\label{sec:encode}
\begin{figure}[!t]
\centering
\subfigure[Our HIN schema]{
    \includegraphics[scale=0.33]{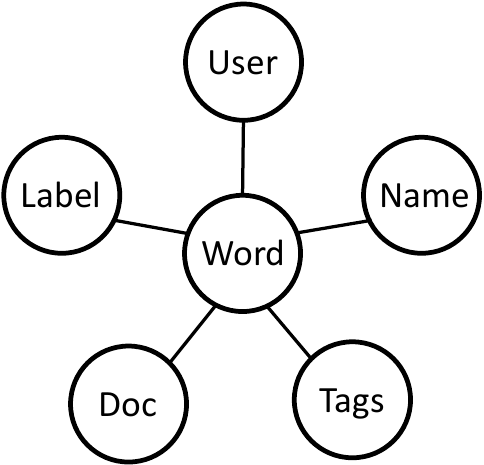}}
  \hspace{-1ex}
  \subfigure[A sample ego network under the schema]{
    \includegraphics[scale=0.33]{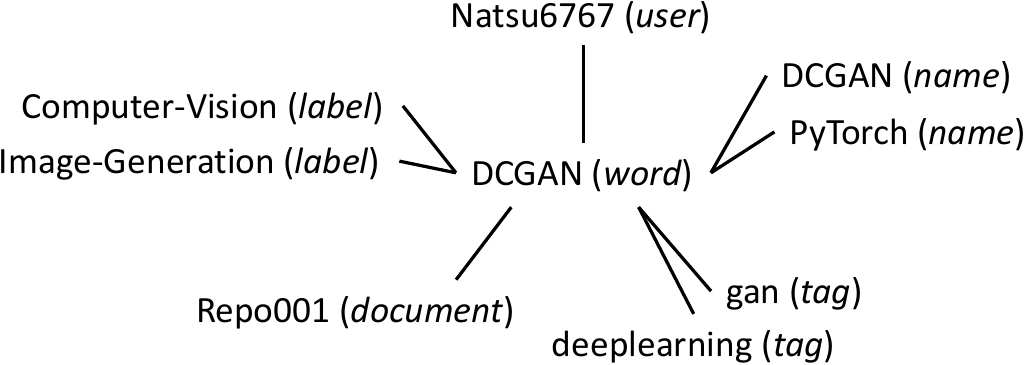}}
\caption{Our HIN schema and a sample network under the schema. The five edge types characterize different kinds of second-order proximity between words.} \label{fig:hin}
\vspace{-1em}
\end{figure}

\noindent \textbf{HIN Construction.} 
The first step of our model is to construct an HIN that can capture all the interactions between different types of information regarding GitHub repositories. 
We include six types of nodes: words ($W$), documents ($D$), users ($U$), tags ($T$), tokens segmented from repository names ($N$) and labels ($L$). There is a one-to-one mapping between documents ($D$) and repositories ($R$), thus document nodes may also serve as a representation of its corresponding repository in the network.  
Since the goal of this module is to learn accurate word representations for the subsequent classification step, we adopt a \textit{word-centric star schema} \cite{sun2009ranking,tang2015pte}. The schema is shown in Fig. \ref{fig:hin}(a). We use a sample ego network of the word ``\textit{DCGAN}'' to help illustrate our schema (Fig. \ref{fig:hin}(b)). The word vocabulary is the union of words present in the documents, tags, segmented repository names and user-provided keywords.

Following the star schema, we then have 5 types of edges in the HIN that represent 5 types of word co-occurrences:  

(1) $W$--$D$. The \textit{word-document} edges describe document-level co-occurrences, where the edge weight between word $w_i$ and document $d_j$ indicates the number of times $w_i$ appears in $d_j$ (i.e., term frequency, or $tf(w_i, d_j)$). From the perspective of \textit{second-order proximity} \cite{tang2015line}, $W$--$D$ edges reflect the fact that two words tend to have similar semantics when they appear in the same repository's document. 

(2) $W$--$U$. 
We add an edge between a word and a user node if the user is the owner of a repository that contains the word in its document field. 
The edge weight between word $w_i$ and user $u_j$ is the sum of the term frequency of the word $w$ in each of the user's repositories: 
$$
\sum_{k:{\rm\ document\ }d_k{\rm \ belongs\ to\ user\ }u_j}tf(w_i, d_k).
$$

(3) $W$--$T$. The \textit{word-tag} relations encode tag-level word co-occurrences. The edge weight between word $w_i$ and tag $t_j$ is 
$$
\sum_{k:{\rm\ document\ }d_k{\rm \ has\ tag\ }t_j}tf(w_i, d_k).
$$

(4) $W$--$N$. We segment the repository name using ``-'', ``\_'' and whitespace as separators. For example, we obtain two tokens ``\textit{DCGAN}'' and ``\textit{PyTorch}'' by segmenting the repository name ``\textit{DCGAN-PyTorch}'' in Fig. \ref{fig:repo}. The edge weight between word $w_i$ and name token $n_j$ also defined through  term frequency:  
$$
\sum_{k:{\rm\ document\ }d_k{\rm \ has\ name \ token\ }n_j}tf(w_i, d_k).
$$

(5) $W$--$L$. The \textit{word-label} relations describe category-level word co-occurrences. Only user-provided keywords will be linked with label nodes and its parents. For example, if we select ``\textit{DCGAN}'' as the keyword of a (leaf) category ``\$\textsc{Image-Generation}'', ``\textit{DCGAN}'' will have links to ``\$\textsc{Image-Generation}'' and all of its parent categories (e.g., ``\$\textsc{Computer-Vision}'').

\vspace{1mm}

\noindent \textbf{HIN Embedding.} Once we have an HIN, we proceed to learn representations for nodes in the network. Then the embedding vectors of word nodes can be applied for repository topic classification. 

There are many popular choices for network representation learning on HINs, such as \textsc{metapath2vec} \cite{dong2017metapath2vec} and \textsc{hin2vec} \cite{fu2017hin2vec}. We adopt \textsc{ESim} \cite{shang2016meta} as our HIN embedding algorithm as it achieves the best performance in our task. (We will validate this choice in Section \ref{sec:hin}.) 
Random walk based methods \cite{perozzi2014deepwalk,grover2016node2vec} have enjoyed  success in learning node embeddings on homogeneous graphs in a scalable manner. 
\textsc{ESim} adapts the idea for use on HINs, and restricts the random walk under guidance of user-specified meta-paths. In \textsc{HiGitClass}, we choose $W$--$D$--$W$, $W$--$U$--$W$, $W$--$T$--$W$, $W$--$N$--$W$ and $W$--$L$--$W$ as our meta-paths, modeling the five different types of second-order proximity between words.

Following the selected meta-paths, we can sample a large number of meta-path instances in our HIN (e.g., $W$--$D$--$W$ is a valid node sequence, while $D$--$W$--$D$ is not). 
Given a meta-path $\mathcal{M}$ and its corresponding node sequence $\mathcal{P} = u_1$--$u_2$--...--$u_l$, we assume that the probability of observing a path given a meta-path constraint follows that of a first-order Markov chain: 
\begin{equation}
    \Pr(\mathcal{P}|\mathcal{M}) = \Pr(u_1|\mathcal{M})\prod_{i=1}^{l-1}\Pr(u_{i+1}|u_i, \mathcal{M}), \notag
\end{equation}
where
\begin{equation}
    \Pr(v|u,\mathcal{M}) = \frac{\exp(f(u,v,\mathcal{M}))}{\sum_{v'\in V}\exp(f(u,v',\mathcal{M})) }
    \label{nodeprob}
\end{equation}
and
\begin{equation}
    f(u,v,\mathcal{M}) = \mu_{\mathcal{M}} + p_{\mathcal{M}}^Te_u + q_{\mathcal{M}}^Te_v + e_u^Te_v. \notag
\end{equation}
Here, $\mu_{\mathcal{M}}$ is the global bias of meta-path $\mathcal{M}$. $p_{\mathcal{M}}$ and $q_{\mathcal{M}}$ are $d$-dimensional local bias of $\mathcal{M}$. $e_u$ and $e_v$ are $d$-dimensional embedding vectors of nodes $u$ and $v$, respectively. $e_u$, $e_v$, $p_{\mathcal{M}}$, $q_{\mathcal{M}}$ and $\mu_{\mathcal{M}}$ can be learned through maximizing the likelihood. 

However, the denominator in Equation (\ref{nodeprob}) requires summing over all nodes, which is very computationally expensive given the large network size. In our actual computation, we estimate this term through negative sampling \cite{mikolov2013distributed}. 
\begin{equation}
    \Pr(v|u,\mathcal{M}) = \frac{\exp(f(u,v,\mathcal{M}))}{\sum_{v'\in V^-}\exp(f(u,v',\mathcal{M}))+ \exp(f(u,v,\mathcal{M}))}, \notag
\end{equation}
where $V^-$ is the set of nodes that serve as negative samples.

\subsection{Keyword Enrichment}
\label{sec:enrich}
Since we only ask users to provide \textit{one} keyword for each category, in case of scarcity and bias, we devise a keyword enrichment module to automatically expand the single keyword $w_{i0}$ the a keyword set $\mathcal{K}_i = \{w_{i0}, w_{i1}, ..., w_{iK_i}\}$ so as to better capture the semantics of the category.

From the HIN embedding step, we have obtained the embedding vector $e_w$ for each word $w$. We perform normalization so that all embedding vectors reside on the unit sphere (i.e., $e_w \leftarrow e_w/||e_w||$).
Then the inner product of two embedding vectors $e_{w_1}^T e_{w_2}$ is adopted to characterize the proximity between two words $w_1$ and $w_2$. For each class $C_i$, we add words sharing the highest proximity with $w_{i0}$ into its enriched keyword set. Meanwhile, to create a clear separation boundary between categories, we require $\mathcal{K}_1,...,\mathcal{K}_\mathcal{L}$ to be \textit{mutually exclusive}. Therefore, the expansion process terminates when any two of the keyword sets tend to intersect. Algorithm \ref{alg:expan} describes the process.

\begin{algorithm}[!t]
	\caption{\textsc{KeywordEnrich}($w_{10},...,w_{{\mathcal{L}0}}$)}
	\label{alg:expan}
	\begin{algorithmic}[1]
		\STATE $\mathcal{K}_i = \{w_{i0}\}$, $i=1,...,\mathcal{L}$
		\STATE $w_{i,last} = w_{i0}$, $i=1,...,\mathcal{L}$
		\WHILE{$\mathcal{K}_i \cap \mathcal{K}_j =\emptyset$ $(\forall i,j)$}
		\FOR {$i=1$ to $\mathcal{L}$}
		    \STATE $w_{i,last} = \arg\max_{w \notin \mathcal{K}_i}e_{w_{0i}}^Te_w$
		    \STATE $\mathcal{K}_i = \mathcal{K}_i \cup \{w_{i,last}\}$
		\ENDFOR
		\ENDWHILE
		\STATE $\mathcal{K}_i = \mathcal{K}_i / \{w_{i,last}\}$, $i=1,...,\mathcal{L}$ \ \ \ //Remove the last added keyword to keep mutual exclusivity
		\STATE output $\mathcal{K}_1,...,\mathcal{K}_\mathcal{L}$
	\end{algorithmic}
\end{algorithm}

Note that on a unit sphere, the inner product is a reverse measure of the spherical distance between two points. Therefore, we are essentially expanding the keyword set with the nearest neighbors of the given keyword. The termination condition is that two ``neighborhoods'' have overlaps.

\subsection{Topic Modeling and Pseudo Document Generation}
To leverage keywords for classification, we face two problems: 
(1) a typical classifier needs labeled repositories as input; (2) although the keyword sets have been enriched, a classifier will likely be overfitted if it is trained solely on these keywords. To tackle these issues,
we assume we can generate a training document $\tilde{d}$ for class $C_i$ given $\mathcal{K}_i$ through the following process: 
\begin{equation}
    q(\tilde{d}| C) = q(\tilde{d}|\Theta_i) p(\Theta_i | \mathcal{K}_i).  \notag
\end{equation}
Here $q(\cdot|\Theta_i)$ is the topic distribution of $C_i$ parameterized by $\Theta_i$, with which we can ``smooth'' the small set of keywords $\mathcal{K}_i$ to a continuous distribution. For simplicity, we adopt a ``bag-of-words'' model for the generated documents, so $ q(\tilde{d}|\Theta_i) = \prod_{i=0}^{|\tilde{d}|} q (w_i|\Theta_i) $. Then we draw samples of words from $q (\cdot|\Theta_i)$  to form a pseudo document following the technique proposed in \cite{meng2018weakly}. 

\vspace{1mm}

\noindent \textbf{Spherical Topic Modeling.} 
Given the normalized embeddings, we characterize the word distribution for each category using a mixture of von Mises-Fisher (vMF) distributions \cite{banerjee2005clustering,gopal2014mises}. To be specific, the probability to generate keyword $w$ from category $C_i$ is defined as
$$
q(w|C_i) = \sum_{j=1}^m \alpha_j f(e_w|\mu_j,\kappa_j) = \sum_{j=1}^m \alpha_j c_p(\kappa_j)\exp(\kappa_j\mu_j^Te_w),
$$
where $f(e_w|\mu_j,\kappa_j)$, as a vMF distribution, is the $j$-th component in the mixture with a weight $\alpha_j$. The vMF distribution can be interpreted as a normal distribution confined to a unit sphere. It has two parameters: the mean direction vector $\mu_i$ and the concentration parameter $\kappa_i$. The keyword embeddings concentrate around $\mu_i$, and are more concentrated if $\kappa_i$ is large. $c_p(\kappa_i)$ is a normalization constant.

Following \cite{meng2019weakly}, we choose the number of vMF components differently for leaf and internal categories: (1) For a \textit{leaf} category $C_j$, the number of components $m$ is set to 1 and the mixture model degenerates to a single vMF distribution. (2) For an \textit{internal} category $C_j$, we set the number of components to be the number of $C_j$'s children in the label hierarchy.

Given the enriched keyword set $\mathcal{K}_j$, we can derive $\mu_j$ and $\kappa_j$ using Expectation Maximization (EM)\cite{banerjee2005clustering}. Recall that the keywords of an internal category are aggregated from its children categories. In practice, we use the approximation procedure based on Newton's method \cite{banerjee2005clustering} to derive $\kappa_j$.


\vspace{1mm}

\noindent \textbf{Pseudo Document Generation.} 
\label{sec:gen}
To generate a pseudo document $\tilde{d}$ for $C_j$, we first sample a document vector $e_{\tilde{d}}$ from $f(\cdot|C_j)$. Then we build a local vocabulary $V_{\tilde{d}}$ that contains top-$\tau$ words similar with $\tilde{d}$ in the embedding space. ($\tau=50$ in our model.) Given $V_{\tilde{d}}$, we repeatedly generate a number of words from a background distribution with probability $\beta$ and from the document-specific distribution with probability $1-\beta$. Formally,
\begin{equation}
\Pr(w|\tilde{d})=
\begin{cases}
\beta p_B(w), &w \notin V_{\tilde{d}}\\
\beta p_B(w)+(1-\beta)\frac{\exp(e_w^Te_{\tilde{d}})}{\sum_{w'\in V_{\tilde{d}}}\exp(e_{w'}^Te_{\tilde{d}})}, &w \in V_{\tilde{d}}  \notag
\end{cases}
\end{equation}
where $p_B(w)$ is the background distribution (i.e., word distribution in the entire corpus).

\begin{figure*}[t]
\centering
\subfigure[\textsc{Machine-Learning}]{
\includegraphics[width=0.45\textwidth]{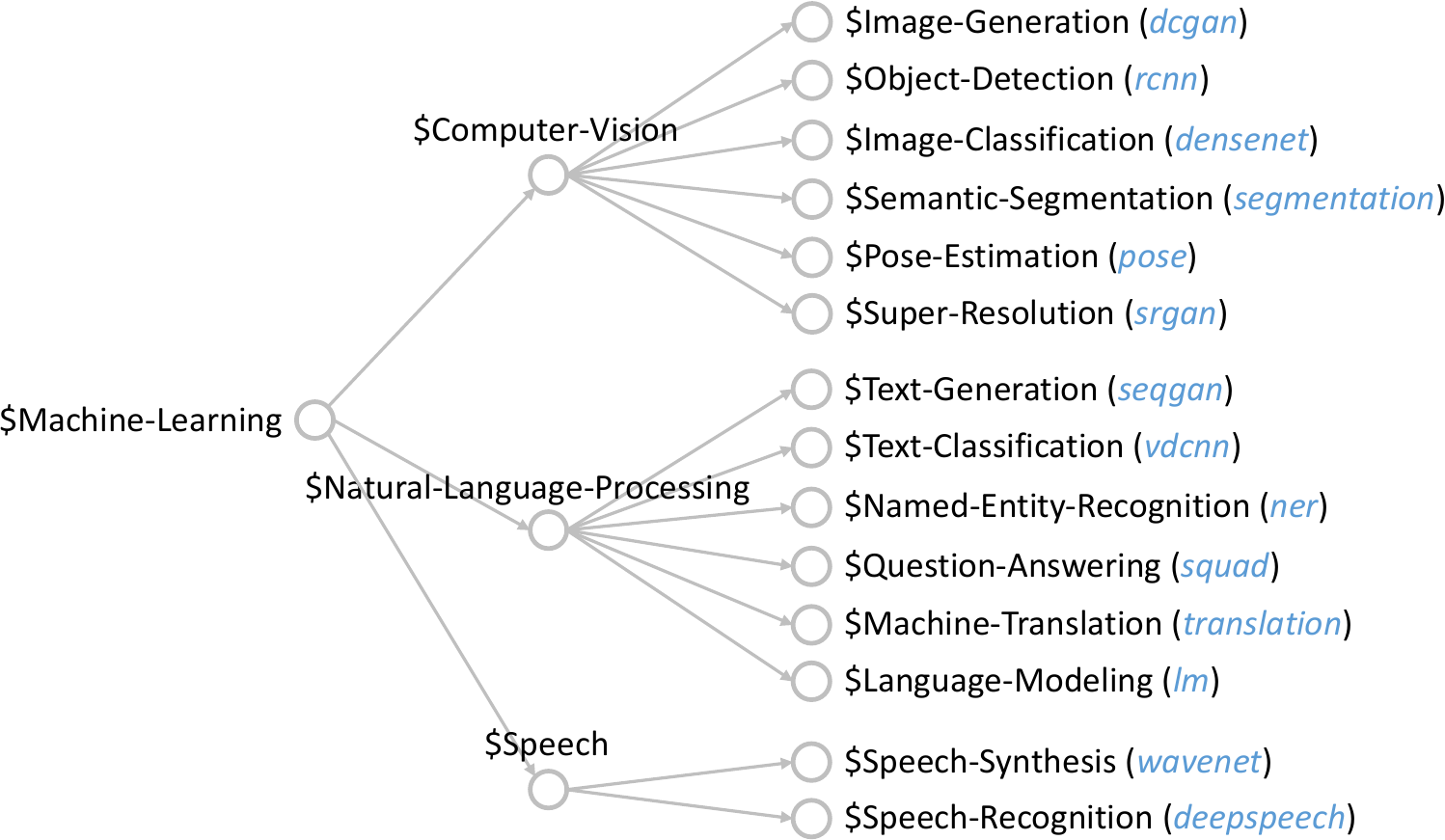}}
\hspace{1ex}
\subfigure[\textsc{Bioinformatics}]{
\includegraphics[width=0.45\textwidth]{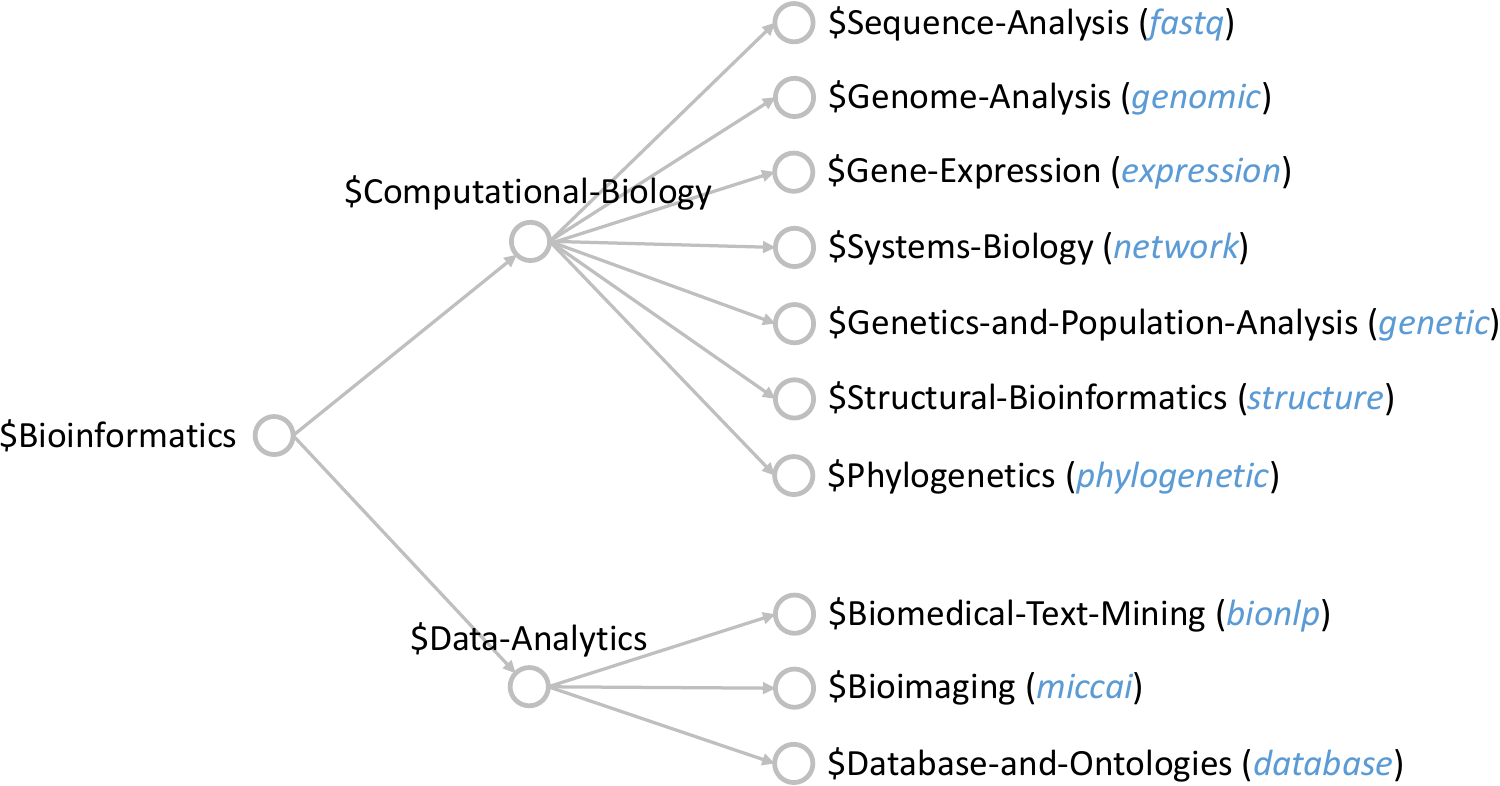}}
\caption{Label hierarchy and provided keywords (in blue) on the two datasets.}
\label{fig:tax}
\vspace{-1em}
\end{figure*}

Here we generate the pseudo document in two steps: first sampling the document vector and then sampling words from a mixture of the document language model and a background language model. 
Compared to directly sampling words from $f(\cdot|C_j)$, the two-step process ensures better coverage of the class distribution. In direct sampling, with high probability we will include words that are close to the centroid of the topic $C_j$. The classifier may learn to ignore all other words and use only these words to determine the predicted class. 
By first sampling the document vector, we would like to lead the classifier to learn that all documents that fall within the topic distribution belong to the same class. 

The synthesized pseudo documents are then used as training data for a classifier. In \textsc{HiGitClass}, we adopt convolutional neural networks (\textsc{CNN}) \cite{kim2014convolutional} for the classification task. One can refer to \cite{kim2014convolutional,meng2018weakly} for more details of the network architecture. The embedding vectors of word nodes obtained by \textsc{ESim} in the previous HIN module are used as pre-trained embeddings.

Recall the process of generating pseudo documents, if we evenly split the fraction of the background distribution into the $m$ children categories of $C_j$, the ``true'' label distribution (an $m$-dimensional vector) of a pseudo document $\tilde{d}$ can be defined as
\begin{equation}
{\rm label}(\tilde{d})_i = 
\begin{cases}
(1 - \beta) + \beta/m, &\tilde{d} \text{ is generated from child }i\\
\beta/m. &\text{otherwise} \notag
\end{cases}
\end{equation}
Since our label is a distribution instead of a one-hot vector, 
we compute the loss as the KL divergence between the output label distribution and the pseudo label. 


%% file: 4-exp.tex
\section{Experiments}
We aim to answer two questions in our experiments. First, does \textsc{HiGitClass} achieve supreme performance in comparison with various baselines (Section \ref{sec:comp})? Second, we propose three key modules in \textsc{HiGitClass}. How do they contribute to the overall performance? (The effects of these three modules will be explored one by one in Sections \ref{sec:hin}, \ref{sec:key} and \ref{sec:doc}).

\subsection{Experimental Setup}
\label{sec:setup}
\noindent \textbf{Datasets.}
We collect two datasets of GitHub repositories covering different domains.\footnote{Our code and datasets are available at \texttt{\url{https://github.com/yuzhimanhua/HiGitClass}}.} Their statistics are summarized in Table \ref{tab:data}.
\begin{table}[H]
\centering
\footnotesize
\caption{Dataset statistics.}
\vspace{-0.5em}
\begin{tabular}{ccc}
\hline
Dataset          & \#Repos & \#Classes (Level 1 + Level 2)  \\ \hline
\textsc{Machine-Learning} & 1,596   & 3 + 14                                                                            \\
\textsc{Bioinformatics}   & 876     & 2 + 10                                                                           \\ \hline
\end{tabular}
\label{tab:data}
\end{table}

\begin{itemize}
    \item \textsc{Machine-Learning.} This dataset is collected by the Paper With Code project\footnote{https://paperswithcode.com/media/about/evaluation-tables.json.gz}. It contains a list of GitHub repositories implementing state-of-the-art algorithms of various machine learning tasks, where the tasks are organized as a taxonomy.
  
    \item \textsc{Bioinformatics.} This dataset is extracted from research articles published on four venues \textit{Bioinformatics}, \textit{BioNLP}, \textit{MICCAI} and \textit{Database} from 2014 to 2018. In each article, authors may put a code link, and we extract the links pointing to a GitHub repository. Meanwhile, each article has an issue section, which is viewed as the topic label of the associated repository. 
\end{itemize}

Note that more than 73\% (resp., 78\%) of the repositories in our \textsc{Machine-Learning} (resp., \textsc{Bioinformatics}) dataset have no tags.

\vspace{1mm}

\begin{table*}[]
\centering
\caption{Performance of Compared Algorithms on the \textsc{Machine-Learning} Dataset. HierDataless does not have a standard deviation since it is a deterministic algorithm.}
\footnotesize
\begin{tabular}{c|cc|cc|cc}
\hline
Method       & Level-1 Micro  & Level-1 Macro  & Level-2 Micro  & Level-2 Macro  & Overall Micro  & Overall Macro  \\ \hline
HierSVM \cite{dumais2000hierarchical}     & 54.20 $\pm$  4.53        & 46.58 $\pm$ 3.59         & 27.40  $\pm$  3.55       & 31.99 $\pm$  5.34        & 40.80 $\pm$ 1.20         & 34.57  $\pm$ 3.94        \\
HierDataless \cite{song2014dataless} & 76.63          & 33.44          & 13.72          & 8.80           & 45.18          & 13.15          \\
WeSTClass \cite{meng2018weakly}   & 61.78 $\pm$ 3.90          & 48.00 $\pm$  2.04       & 37.71 $\pm$  2.72      & 34.34  $\pm$   1.70     & 49.75 $\pm$  3.24       & 36.75  $\pm$  1.67      \\
WeSHClass \cite{meng2019weakly}   & 70.69 $\pm$ 2.14       & 52.73 $\pm$  2.18       & 38.08 $\pm$ 2.07         & 33.87  $\pm$ 2.23       & 54.39  $\pm$  2.11      & 37.60 $\pm$ 1.67        \\
PCNB    \cite{xiao2019effi}     & 77.79 $\pm$   1.92      & 62.53  $\pm$ 2.55       & 26.77 $\pm$ 3.89        & 22.85  $\pm$ 1.98       & 52.28 $\pm$  1.92       & 29.85 $\pm$ 1.74        \\
PCEM    \cite{xiao2019effi}     & 77.28  $\pm$  2.00      & 59.27 $\pm$ 2.29        & 22.34     $\pm$ 4.02    & 19.28  $\pm$ 2.25       & 49.81  $\pm$ 2.00       & 26.34  $\pm$ 1.83       \\ \hline
\textsc{HiGitClass w/o HIN}       &   75.28  $\pm$ 6.99          &    60.85  $\pm$ 5.51         & 40.21 $\pm$ 2.91             &  39.77  $\pm$ 2.22           & 57.74    $\pm$ 4.07          &  43.49  $\pm$  2.10          \\
\textsc{HiGitClass w/o Enrich} & 86.48 $\pm$  1.41      & \textbf{72.19 $\pm$  2.53}       & 36.71 $\pm$ 1.13        & 43.75  $\pm$ 2.85       & 61.60 $\pm$ 0.25        & 48.77 $\pm$  1.94        \\
\textsc{HiGitClass w/o Hier} & 57.31 $\pm$  1.63      & 59.30 $\pm$  3.14       & 40.45 $\pm$ 2.51        & 44.41  $\pm$ 2.66       & 48.88 $\pm$ 2.07        & 47.04 $\pm$  1.63       \\
\textsc{HiGitClass}   & \textbf{87.68 $\pm$ 2.23} & 72.06 $\pm$ 4.39 & \textbf{43.93 $\pm$ 2.93} & \textbf{45.97 $\pm$ 2.29} & \textbf{65.81 $\pm$ 3.55} & \textbf{50.57 $\pm$ 2.98} \\
\hline
\end{tabular}
\label{tab:aif1}
\end{table*}

\begin{table*}[]
\centering
\caption{Performance of Compared Algorithms on the \textsc{Bioinformatics} Dataset. HierDataless does not have a standard deviation since it is a deterministic algorithm.}
\footnotesize
\begin{tabular}{c|cc|cc|cc}
\hline
Method       & Level-1 Micro & Level-1 Macro  & Level-2 Micro  & Level-2 Macro  & Overall Micro  & Overall Macro  \\ \hline
HierSVM \cite{dumais2000hierarchical}     & 80.39 $\pm$ 2.72        & 70.16 $\pm$ 6.05        & 13.49 $\pm$ 10.2        & 10.04  $\pm$ 9.07       & 46.94  $\pm$ 4.66       & 20.06  $\pm$ 7.12       \\
HierDataless \cite{song2014dataless} & 81.39          & \textbf{78.75}          & 39.27          & 36.57          & 60.33          & 43.60          \\
WeSTClass \cite{meng2018weakly}    & 61.78 $\pm$ 5.75        & 52.73 $\pm$ 4.86        & 20.39 $\pm$  3.48       & 17.52 $\pm$ 2.59        & 41.09 $\pm$  4.28       & 23.91 $\pm$ 2.82        \\
WeSHClass \cite{meng2019weakly}    & 63.17 $\pm$ 3.56       & 59.65 $\pm$ 3.51         & 26.44 $\pm$ 1.33         & 24.94 $\pm$  0.98       & 44.80 $\pm$  2.26       & 30.72 $\pm$ 1.30        \\
PCNB  \cite{xiao2019effi}       & 77.03 $\pm$ 2.89        & 59.43 $\pm$ 3.51        & 31.77    $\pm$  3.80    & 22.90  $\pm$  4.84      & 54.40  $\pm$ 2.89       & 28.99  $\pm$   4.82     \\
PCEM  \cite{xiao2019effi}       & 78.51 $\pm$ 3.06        & 61.99 $\pm$ 4.18         & 32.80   $\pm$ 2.88      & 18.93 $\pm$ 5.28        & 55.66 $\pm$  3.06       & 26.11 $\pm$ 5.41        \\ \hline
\textsc{HiGitClass w/o HIN}       &   66.21 $\pm$  8.33          &  64.39 $\pm$ 7.00            &   31.87 $\pm$ 3.65           &   30.47 $\pm$  3.15          &  49.04  $\pm$   5.75         &    36.13 $\pm$  3.52         \\
\textsc{HiGitClass w/o Enrich} & 54.55 $\pm$ 10.0        & 53.57 $\pm$ 9.15        & 22.95 $\pm$ 1.46        & 23.18 $\pm$ 1.89        & 38.74  $\pm$  4.30      & 28.24 $\pm$ 0.87        \\
\textsc{HiGitClass w/o Hier} & 78.79 $\pm$  2.57      & 73.71 $\pm$  2.28       & 39.42 $\pm$  4.47        & 41.58  $\pm$ 2.63       & 59.10 $\pm$ 3.51        & 46.94 $\pm$  2.48        \\
\textsc{HiGitClass}   & \textbf{81.71 $\pm$ 3.95} & 77.11 $\pm$ 3.89  & \textbf{42.44 $\pm$ 8.46} & \textbf{41.67 $\pm$ 8.04} & \textbf{62.08 $\pm$ 6.11} & \textbf{47.57 $\pm$ 7.34} \\
\hline
\end{tabular}
\label{tab:biof1}
\end{table*}

\noindent \textbf{Baselines.} We evaluate the performance of \textsc{HiGitClass} against the following hierarchical classification algorithms:

\begin{itemize}
	\item \textbf{HierSVM} \cite{dumais2000hierarchical} decomposes the training tasks according to the label taxonomy, where each local SVM is trained to distinguish sibling categories that share the same parent node.\footnote{https://github.com/globality-corp/sklearn-hierarchical-classification}
	\item \textbf{HierDataless} \cite{song2014dataless} embeds both class labels and documents in a semantic space using Explicit Semantic Analysis on Wikipedia articles, and assigns the nearest label to each document in the semantic space.\footnote{https://github.com/yqsong/DatalessClassification} Note that HierDataless uses Wikipedia as \textbf{external knowledge} in classification, whereas other baselines and \textsc{HiGitClass} solely reply on user-provided data.
	\item \textbf{WeSTClass} \cite{meng2018weakly} first generates pseudo documents and then trains a \textsc{CNN} based on the synthesized training data.\footnote{https://github.com/yumeng5/WeSTClass}
	\item \textbf{WeSHClass} \cite{meng2019weakly} leverages a language model to generate synthesized data for pre-training and then iteratively refines the global hierarchical model on labeled documents.\footnote{https://github.com/yumeng5/WeSHClass}
    \item \textbf{PCNB} \cite{xiao2019effi} utilizes a path-generated probabilistic framework on the label hierarchy and trains a path-cost sensitive naive Bayes classifier.\footnote{ https://github.com/HKUST-KnowComp/PathPredictionForTextClassifica-tion}
	\item \textbf{PCEM} \cite{xiao2019effi} makes use of the unlabeled data to ameliorate the path-cost sensitive classifier and applies an EM technique for semi-supervised learning.
\end{itemize}

Note that HierSVM, PCNB and PCEM can only take document-level supervision (i.e., labeled repositories). To align the experimental settings, we first label all the repositories using TFIDF scores by treating the keyword set of each class as a query. Then, we select top-ranked repositories per class as the supervision to train HierSVM, PCNB and PCEM. Since the baselines are all text classification approaches, we append the information of user, tags and repository name to the end of the document for each repository so that the baselines can exploit these signals.

Besides the baselines, we also include the following three ablation versions of \textsc{HiGitClass} into comparison.
\begin{itemize}
	\item \textbf{w/o HIN} skips the HIN embedding module and relies on word2vec \cite{mikolov2013distributed} to generate word embeddings for the following steps. 
	\item \textbf{w/o Enrich} skips the keyword enrichment module and directly uses one single keyword in spherical topic modeling.
	\item \textbf{w/o Hier} directly classifies all repositories to the leaf layer and then assigns internal labels to each repository according to its leaf category.
\end{itemize}

\vspace{1mm}

\noindent \textbf{Evaluation Metrics.} We use F1 scores to evaluate the performance of all methods. Denote $TP_i$, $FP_i$ and, $FN_i$ as the instance numbers of true-positive, false-positive and false negative for category $C_i$. Let $\mathcal{T}_1$ (resp., $\mathcal{T}_2$) be the set of all Level-1 (resp., Level-2/leaf) categories. The Level-1 Micro-F1 is defined as $\frac{2PR}{P+R}$, where $P=\frac{\sum_{C_i \in \mathcal{T}_1} TP_i}{\sum_{C_i \in \mathcal{T}_1} (TP_i+FP_i)}$ and $R=\frac{\sum_{C_i \in \mathcal{T}_1} TP_i}{\sum_{C_i \in \mathcal{T}_1} (TP_i+FN_i)}$. The Level-1 Macro-F1 is defined as $\frac{1}{|\mathcal{T}_1|}\sum_{C_i \in \mathcal{T}_1}\frac{2P_iR_i}{P_i+R_i}$, where $P_i = \frac{TP_i}{TP_i+FP_i}$ and $R_i = \frac{TP_i}{TP_i+FN_i}$. Accordingly, Level-2 Micro/Macro-F1 and Overall Micro/Macro-F1 can be defined on $\mathcal{T}_2$ and $\mathcal{T}_1 \cup \mathcal{T}_2$.

\subsection{Performance Comparison with Baselines}
\label{sec:comp}

Tables \ref{tab:aif1} and \ref{tab:biof1} demonstrate the performance of compared methods on two datasets. We repeat each experiment 5 times (except HierDataless, which is a deterministic algorithm) with the mean and
standard deviation reported.

As we can observe from Tables \ref{tab:aif1} and \ref{tab:biof1}, on both datasets, a significant improvement is achieved by \textsc{HiGitClass} compared to the baselines. On \textsc{Machine-Learning}, \textsc{HiGitClass} notably outperforms the second best approach by 22.1\% on average. On \textsc{Bioinformatics}, the only metric in terms of which \textsc{HiGitClass} does not perform the best is Level-1 Macro-F1, and the main opponent of \textsc{HiGitClass} is HierDataless. As mentioned above, HierDataless incorporates Wikipedia articles as its external knowledge. When user-provided keywords can be linked to Wikipedia (e.g., ``\textit{genomic}'', ``\textit{genetic}'' and ``\textit{phylogenetic}'' in \textsc{Bioinformatics}), HierDataless can exploit the external information well. However, when the keywords cannot be wikified (e.g., names of new deep learning algorithms such as ``\textit{dcgan}'', ``\textit{rcnn}'' and ``\textit{densenet}'' in \textsc{Machine-Learning}), the help from Wikipedia is limited. In fact, on \textsc{Machine-Learning}, HierDataless performs poorly. Note that on GitHub, it is common that names of recent algorithms or frameworks are provided as keywords.

Besides outperforming baseline approaches, \textsc{HiGitClass} shows a consistent and evident improvement against three ablation versions. The average boost of the HIN module over the six metrics is 15.0\% (resp., 28.6\%) on the \textsc{Machine-Learning} (resp., \textsc{Bioinformatics}) dataset, indicating the importance of encoding multi-modal signals on GitHub. The Level-1 F1 scores of \textsc{HiGitClass w/o Enrich} is close to the full model on \textsc{Machine-Learning}, but the keyword enrichment module demonstrates its power when we go deeper. This finding is aligned with the fact that the topic distributions of coarse-grained categories are naturally distant from each other on the sphere. Therefore, one keyword per category may be enough to estimate the distributions approximately. However, when we aim at fine-grained classification, the topic distributions become closer, and the inference process may be easily interfered by a biased keyword. \textsc{HiGitClass w/o Hier} performs poorly on \textsc{Machine-Learning}, which highlights the importance of utilizing the label hierarchy during the training process. The same phenomenon occurs in the comparison between WeSTClass and WeSHClass.

\subsection{Effect of HIN Construction and Embedding}
\label{sec:hin}
\begin{figure}[t]
\centering
  \subfigure[\textsc{Machine-Learning}, Micro]{
    \includegraphics[width=0.23\textwidth]{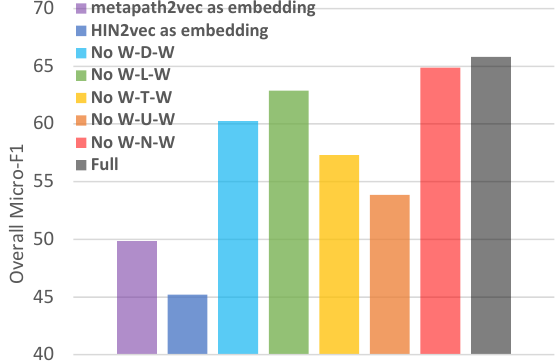}}
  \hspace{0ex}
  \subfigure[\textsc{Bioinformatics}, Micro]{
    \includegraphics[width=0.23\textwidth]{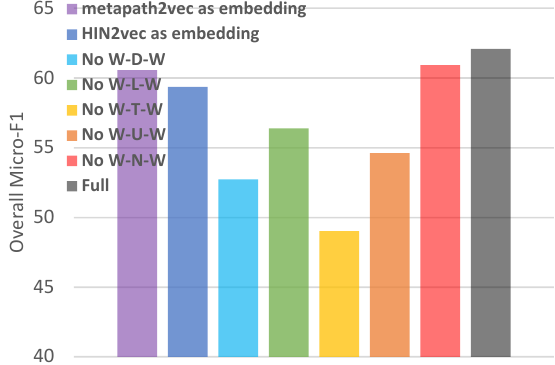}}
  \vspace{0mm}
  \subfigure[\textsc{Machine-Learning}, Macro]{
    \includegraphics[width=0.23\textwidth]{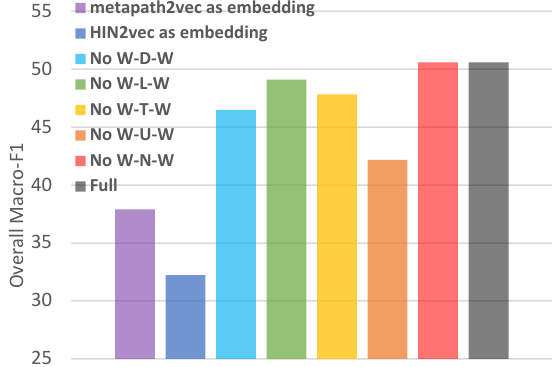}}
  \hspace{0ex}
  \subfigure[\textsc{Bioinformatics}, Macro]{
    \includegraphics[width=0.23\textwidth]{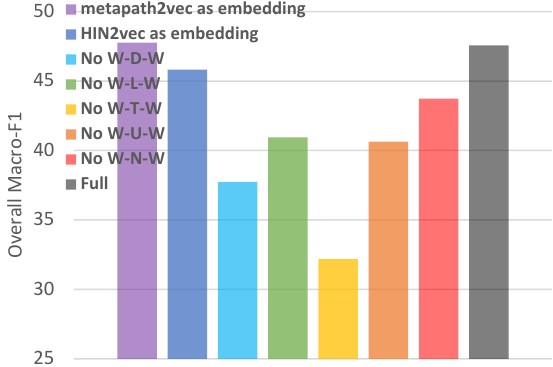}}
  \caption{Performance of algorithms with different HIN modules. }
  \label{fig:embedding}
  \vspace{-0.5em}
\end{figure}

We have demonstrated the contribution of our HIN module by comparing \textsc{HiGitClass} and \textsc{HiGitClass w/o HIN}. To explore the effectiveness of HIN construction and embedding in a more detailed way, we perform an ablation study by changing one ``factor'' in the HIN and fixing all the other modules in \textsc{HiGitClass}. To be specific, our HIN has five types of edges, each of which corresponds to a meta-path. We consider five ablation versions (\textbf{No W-D-W}, \textbf{No W-U-W}, \textbf{No W-T-W}, \textbf{No W-N-W} and \textbf{No W-L-W}). Each version ignores one edge type/meta-path. Moreover, given the complete HIN, we consider to use two popular approaches, \textsc{metapath2vec} \cite{dong2017metapath2vec} and \textsc{hin2vec} \cite{fu2017hin2vec}, as our embedding technique, which generates two variants \textbf{metapath2vec as embedding} and \textbf{HIN2vec as embedding}. Fig. \ref{fig:embedding} shows the performance of these variants and our \textbf{Full} model.

We have the following findings from Fig. \ref{fig:embedding}. First, our \textsc{Full} model outperforms the five ablation models ignoring different edge types, indicating that each meta-path (as well as each node type incorporated in the HIN construction step) plays a positive role in classification. Second, our \textsc{Full} model outperforms \textsc{metapath2vec as embedding} and \textsc{hin2vec as embedding} in most cases, which validates our choice of using \textsc{ESim} as the embedding technique. The possible reason that \textsc{ESim} is more suitable for our task may be that we have a simple word-centric star schema and a clear goal of embedding word nodes. Therefore, the choices of meta-paths can be explicitly specified (i.e., $W$--?--$W$) and do not need to be inferred from data (as \textsc{hin2vec} does).
Third, among the five ablation models ignoring edge types, \textsc{No W-U-W} performs the worst on \textsc{Machine-Learning}, which means $W$-$U$ edges (i.e., the user information) contribute the most in repository classification. Meanwhile, on \textsc{Bioinformatics}, $W$-$T$ edges have the largest offering. This can be explained by the following statistics: in the \textsc{Machine-Learning} dataset, 348 pairs of repositories share the same user, out of which 217 (62\%) have the same \textit{leaf} label; in the \textsc{Bioinformatics} dataset, there are 356 pairs of repositories having at least two overlapping tags, among which 221 (62\%) belong to the same \textit{leaf} category. Fourth, $W$-$D$ edges also contribute a lot to the performance. This observation is aligned with the results in \cite{tang2015pte}, where document-level word co-occurrences play a crucial role in text classification.   

\subsection{Effect of Keyword Enrichment}
\label{sec:key}
\begin{table*}[]
\small
\centering
\caption{Entity enrichment results on the two datasets. Four leaf categories are shown for each dataset.}
\scalebox{0.93}{
\begin{tabular}{c|cccc}
\hline
Class                              & \textsc{\$Semantic-Segmentation} & \textsc{\$Pose-Estimation} & \textsc{\$Named-Entity-Recognition} & \textsc{\$Question-Answering}  \\ \hline
Keyword                   & segmentation          & pose            & ner                      & squad                      \\ \hline
\multirow{5}{*}{\begin{tabular}[c]{@{}c@{}}Enriched\\ Keywords\end{tabular}} & semantic              & estimation      & entity                   & question                        \\
                                   & papandreou            & person          & tagger                   & answering                     \\
                                   & scene                 & human           & lample                   & bidaf                          \\
                                   & pixel                 & mpii            & -                        & -                          \\
                                   & segment               & 3d              & -                        & -                             \\ \hline
\hline
Class                              & \textsc{\$Gene-Expression} & \textsc{\$Genetics-and-Population} & \textsc{\$Structural-Bioinformatics} & \textsc{\$Phylogenetics}  \\ \hline
Keyword                   & expression      & genetic                 & structure                 & phylogenetic                 \\ \hline
\multirow{5}{*}{\begin{tabular}[c]{@{}c@{}}Enriched\\ Keywords\end{tabular}} & gene            & traits                  & protein                   & trees                         \\
                                   & genes           & trait                   & pdb                       & newick                              \\
                                   & rna             & markers                 & residues                  & phylogenetics                      \\
                                   & cell            & phenotypes              & pymol                     & phylogenies                         \\
                                   & isoform         & associations            & residue                   & evolution                          \\ \hline
\end{tabular}
}
\label{tab:keyword}
\end{table*}
Quantitatively, the keyword enrichment module has a positive contribution to the whole framework according to previous experiments. 
We now show its effect qualitatively. Table \ref{tab:keyword} lists top-5 words selected by \textsc{HiGitClass} during keyword enrichment. Besides topic-indicating words (e.g., ``\textit{tagger}'', ``\textit{trait}'', ``\textit{trees}'', etc.), popular algorithm/tool names (e.g., ``\textit{bidaf}'' and ``\textit{pymol}''), dataset names (e.g., ``\textit{mpii}'', ``\textit{pdb}'') and author names (e.g., ``\textit{papandreou}'' and ``\textit{lample}'') are also included in the expanded keyword set. Note that some provided keywords are more or less ambiguous (e.g., ``\textit{segmentation}'' and ``\textit{structure}''), and directly using them for topic modeling may introduce noises. In contrast, the expanded set as a whole can better characterize the semantics of each topic category.

\subsection{Effect of Pseudo Documents}
\label{sec:doc}
\begin{figure}[t]
\centering
  \subfigure[\textsc{Machine-Learning}]{
    \includegraphics[width=0.23\textwidth]{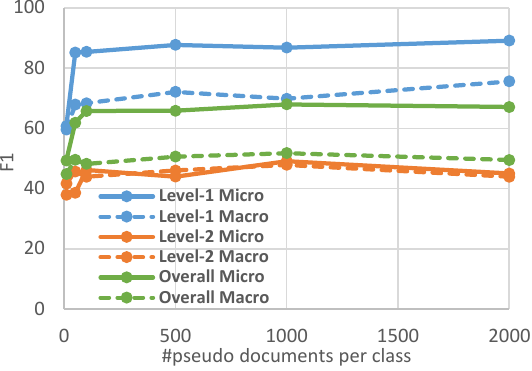}}
  \hspace{0ex}
  \subfigure[\textsc{Bioinformatics}]{
    \includegraphics[width=0.23\textwidth]{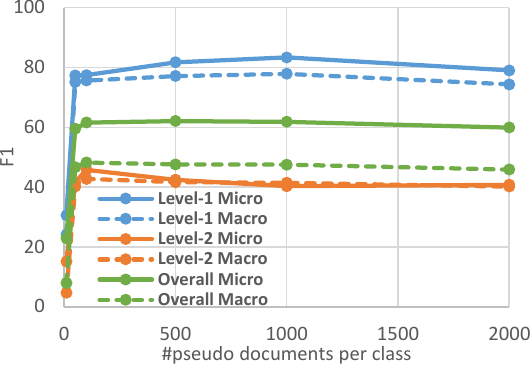}}
  \caption{Performance of \textsc{HiGitClass} with different numbers of pseudo documents.}
  \label{fig:pseudo}
  \vspace{-0.5em}
\end{figure}

In all previous experiments, when we build a classifier for an internal category $C_i$, we generate 500 pseudo documents for each child of $C_i$. What if we use less/more synthesized training data? Intuitively, if the amount of generated pseudo documents is too small, signals in previous modules cannot fully propagate to the training process. On the contrary, if we have too many generated data, the training time will be unnecessarily long. To see whether 500 is a good choice, we plot the performance of \textsc{HiGitClass} with 10, 50, 100, 500, 1000 and 2000 pseudo documents in Fig. \ref{fig:pseudo}. 

On the one side, when the number of pseudo documents is too small (e.g., 10, 50 or 100), information carried in the synthesized training data will be insufficient to train a good classifier. On the other side, when we generate too many pseudo documents (e.g, 1000 or 2000), putting efficiency aside, the performance is not guaranteed to increase. In fact, on both datasets, the F1 scores start to fluctuate when the number of pseudo documents becomes large. In our task, generating 500 to 1000 pseudo documents for each class will strike a good balance.

%% file: 5-related.tex
\section{Related Work}
\noindent \textbf{GitHub Repository Mining.} As a popular code collaboration community, GitHub presents many opportunities for researchers to learn how people write code and design tools to support the process. 
As a result, GitHub data has received attention from both software engineering and social computing researchers. 
Analytic studies \cite{dabbish2012social,tsay2012social,tsay2014influence,kalliamvakou2014promises,russell2018large,ma2017developers} have investigated how user activities (e.g., collaboration, following and watching) affect development practice. Algorithmic studies \cite{zhang2017detecting,sharma2017cataloging,prana2018categorizing,fan2019idev} exploit README files and repository metadata to perform data mining tasks such as similarity search \cite{zhang2017detecting} and clustering \cite{sharma2017cataloging}. 
In this paper, we focus on the task of automatically classifying repositories whereas 
previous works \cite{kalliamvakou2014promises,russell2018large} have relied on human effort to annotate each repository with its topic. 

\vspace{1mm}

\noindent \textbf{HIN Embeddings.}
Many node embeddings techniques have been proposed for HIN, including \cite{shang2016meta,dong2017metapath2vec,fu2017hin2vec,yang2018meta}. 
From the application point of view, typical applications of learned embeddings include node classification \cite{shang2016meta,dong2017metapath2vec,fu2017hin2vec,yang2018meta}, node clustering \cite{dong2017metapath2vec} and link prediction \cite{fu2017hin2vec,yang2018meta}.  
Several studies apply HIN node embeddings into downstream classification tasks, such as malware detection \cite{hou2017hindroid} and medical diagnosis \cite{hosseini2018heteromed}. Different from the fully-supervised settings in \cite{hou2017hindroid,hosseini2018heteromed}, our repository classification task relies on a very small set of guidance. Moreover, most information used in \cite{hou2017hindroid,hosseini2018heteromed} is structured. In contrast, we combine structured information such as user-repository ownership relation with unstructured text for classification. 

\vspace{1mm}

\noindent \textbf{Dataless Text Classification.}
Although deep neural architectures \cite{kim2014convolutional,yang2016hierarchical,Howard2018UniversalLM} demonstrate their advantages in fully-supervised text classification, their requirement of massive training data prohibits them from being adopted in some practical scenarios. Under weakly-supervised or dataless settings, there have been solutions following two directions: \textit{latent variable models} extending topic models (e.g., PLSA and LDA) by incorporating user-provided seed information \cite{lu2008opinion,chen2015dataless,li2016effective,Li2018DatalessDocumentManifold} and  \textit{embedding-based models} deriving vectorized representations for words and documents \cite{chang2008importance,tang2015pte,meng2018weakly}. There are also some work on semi-supervised text classification \cite{Miyato2016AdversarialTM,Sachan2019MixedObjective}, but they require a set of labeled documents instead of keywords. 

\vspace{1mm}

\noindent \textbf{Hierarchical Text Classification.}
Under \textit{fully-supervised} settings, \cite{dumais2000hierarchical} and \cite{liu2005support} first propose to train SVMs to distinguish the children classes of the same parent node. \cite{cai2004hierarchical} further defines hierarchical loss function and applies cost-sensitive learning to generalize SVM learning for hierarchical classification. \cite{peng2018large} proposes a graph-CNN based model to convert text to graph-of-words, on which the graph convolution operations are applied for feature extraction. Under \textit{weakly-supervised} or \textit{dataless} settings, previous approaches include HierDataless \cite{song2014dataless}, WeSHClass \cite{meng2019weakly} and PCNB/PCEM \cite{xiao2019effi}, which have been introduced in Section \ref{sec:setup}. All above mentioned studies focus on text data without additional information. In \textsc{HiGitClass}, we are able to go beyond plain text classification and utilize multi-modal signals.

%% file: 6-conclusion.tex
\section{Conclusions and Future Work}
In this paper, we have studied the problem of 
\textit{keyword-driven hierarchical} classification of GitHub repositories: the end user only needs to provide a label hierarchy and one keyword for each leaf category. Given such scarce supervision, we design \textsc{HiGitClass} with three modules: heterogeneous information network embedding; keyword enrichment; pseudo document generation. Specifically, HIN embeddings take advantage of the multi-modal nature of GitHub repositories; keyword enrichment alleviates the supervision scarcity of each category; pseudo document generation transforms the keywords into documents, enabling the use of powerful neural classifiers.
Through experiments on two repository collections, we show that \textsc{HiGitClass} consistently outperforms all baselines by a large margin, particularly on the lower levels of the hierarchy. 
Further analysis shows that the HIN module contributes the most to the boost in performance and keyword enrichment demonstrates its power deeper down the hierarchy where the differences between classes become more subtle. 

For future work, we would like to explore the possibility of integrating different forms of supervision (e.g., the combination of keywords and labeled repositories). 
The HIN embedding module may also be coupled more tightly with the document classification process by allowing the document classifier's prediction results to propagate along the network. 